\def\year{2020}\relax
\documentclass[letterpaper]{article} 
\usepackage{aaai20}  
\usepackage{times}  
\usepackage{helvet} 
\usepackage{courier}  
\usepackage[hyphens]{url}  
\usepackage{graphicx} 
\def\UrlFont{\rm}  
\usepackage{graphicx}  
\usepackage{epsf}
\usepackage{float}
\usepackage{subfigure}
\usepackage{mathtools}
\usepackage{amsmath}
\usepackage{amssymb}
\usepackage{booktabs}
\usepackage{xcolor}
\usepackage{comment}

\title{Bi-Directional Generation for Unsupervised Domain Adaptation}
\author{Guanglei Yang$^{\dag}$, Haifeng Xia$^{\ddag}$, Mingli Ding$^{\dag}$, Zhengming Ding$^\flat$\thanks{Corresponding author: Zhengming Ding (zd2@iu.edu). This work is done under the supervision of Dr. Zhengming Ding.}\\
$^\dag$School of Instrument science and engineering, Harbin Institute of Technology\\
$^{\ddag}$Department of Electrical \& Computer Engineering, Indiana University-Purdue University Indianapolis\\
$^\flat$Department of Computer, Information and Technology, Indiana University-Purdue University Indianapolis\\
 \{yangguanglei,dingml\}@hit.edu.cn, \{haifxia,zd2\}@iu.edu
}

\begin{document}

\maketitle

\begin{abstract}
Unsupervised domain adaptation facilitates the unlabeled target domain relying on well-established source domain information. 
The conventional methods forcefully reducing the domain discrepancy in the latent space will result in the destruction of intrinsic data structure. 
To balance the mitigation of domain gap and the preservation of the inherent structure, we propose a Bi-Directional Generation domain adaptation model with consistent classifiers interpolating two intermediate domains to bridge source and target domains. Specifically, two cross-domain generators are employed to synthesize one domain conditioned on the other.
The performance of our proposed method can be further enhanced by the consistent classifiers and the cross-domain alignment constraints. 
We also design two classifiers which are jointly optimized to maximize the consistency on target sample prediction. 
Extensive experiments verify that our proposed model outperforms the state-of-the-art on standard cross domain visual benchmarks.
\end{abstract}

\section {Introduction}

Deep learning gains huge success in diverse applications across many fields, such as computer vision, data mining, and natural language processing.
In practical application, it is usually easy to acquire abundant target data. However, the insufficiency and absence of labels is still a challenge \cite{ding2018robust,saito2018maximum} leading to exceedingly time-consuming and expensive manual annotation.
To solve this problem, domain adaptation is introduced and has shown excellent performance. 
It transfers knowledge from external well-labeled source domain to the target domain \cite{wei2016deep,ding2016robust,yan2017mind}.
The theory of domain adaptation is to discover the common latent factors across the source and target domains, and thereby reducing domain mismatch across domains. 
Different domain adaptation methods have been developed, such as feature alignment and domain confusion adaptation \cite{pan2019transferrable}. 

A common practice in domain adaption is to align feature distribution between source and target domains by minimizing domain shift through metrics such as correlation distances \cite{yao2015semi,sun2016return} or maximum mean discrepancy (MMD) \cite{ding2018graph,ding2018deep}.
Among the recent efforts on domain adaption, several works \cite{pan2019transferrable,pinheiro2018unsupervised,Kang_2019_CVPR} aim to minimize the MMD-based distance  of cross-domain in the deep neural network. 
For example, \cite{pan2019transferrable} applies the pairwise reproducing kernel Hilbert space (RKHS) between the prototypes of the same class from different domains to reduce the discrepancy of cross-domain. 
\cite{Kang_2019_CVPR} divides the domain discrepancy into two kinds of class-aware domain discrepancy.
Another branch of domain adaptation is to exploit the domain adversarial training \cite{tzeng2017adversarial,zhang2018fully,saito2018maximum}.
For example, \cite{saito2018maximum} maximizes the discrepancy between two classifiers' outputs to determine target samples without the support of the source. 

Despite the success of previous methods, they suffer from following limitations.
First of all, almost methods apply a shared generator, which means their domain adaption is monodirectional. 
For example, DAN \cite{long2015learning} aligns source domain to target domain.
The features of the target domain are merely simulated instead of positively taking part in domain adaption.
Therefore, such methods cannot fully take advantage of the target data.
Meanwhile, the target data has many unknown factors because it may come from complicated situations.
The boundary among different classes can be not clear, and thereby it hurts the knowledge transfer.
Secondly, most of the previous methods calculate the domain discrepancy at the domain level, but they neglect the class level where the samples come.
Class-agnostic adaptation aligning source and target data at the domain-level is possible to cause sub-optimal solutions.
These solutions are likely to overfit in the source domain, causing that performance in the target domain is not as expectancy.

To address these above issues, we propose a Bi-Directional Generation (BDG) method for unsupervised domain adaptation with dual consistent classifiers. 
The dual generators interpolate two intermediate domains and synthesize more effective data so that there are more samples to train the classifiers. 
Meanwhile, the target samples are labeled by the pre-trained network. 
Apart from  minimizing the adversarial loss on source branch, we employ domain confusion loss in the target branch so that both domain information can be more sufficiently utilized. 
Moreover, we exploit class-wise loss to preserve the class-level semantic information when the generators synthesize two intermediate domains reducing domain discrepancy.
Furthermore, the consistent loss alternatively reduces the classifier discrepancy by minimizing itself directly while the previous methods implement adversarial training in classifiers by discrepancy loss. 
In conclusion, the contributions of our paper are highlighted in three folds as follows:
\begin{itemize}
\item We propose a novel bi-directional cross-domain generation module, which aims to synthesize the intermediate domains conditioned on each domain. 
The augmented samples play as a bridge to reduce the domain discrepancy and preserve the class-level structure during domain adaption process. 
\item We explore dual classifiers to enhance the bi-directional cross-domain generation. 
Furthermore, a consistent loss is developed to improve the prediction performance on unlabeled target samples.
\end{itemize}

\section{Related Works}
In this section, we will briefly introduce two branches of domain adaptation, then highlight the difference of our model.

\subsection{Deep Domain Adaptation}

The aim of domain adaptation is to improve the target learning by using the labeled source knowledge, whose distribution is different from the target domain. 
Spurred by the recent advances in computer vision using deep convolutional neural networks (DCNNs), a number of methods  based  have been proposed for unsupervised domain adaptation.
In particular, one common solution is to guide DCNNs to learn domain feature by minimizing the domain discrepancy with Maximum Mean Discrepancy (MMD) in reproducing kernel Hilbert space (RKHS) \cite{gretton2012kernel}.
MMD is a useful and popular non-parametric metric for the measurement in domain discrepancy between source and target domains. 
\textit{Long et al.} learns more transferable features through multi-kernel MMD loss \cite{long2015learning}. Then, they further extended this work by adding varieties of MMD loss. 
For example, RTN \cite{long2016unsupervised} improves DAN by replacing the multi-kernel MMD loss in DAN with a single tensor-based MMD loss and adds the residual module to AlexNet; JAN \cite{long2017deep} exploits joint maximum mean discrepancy criterion to the loss in order to learn a transfer network. 
Ding et al. explore deep low-rank coding to extract domain-invairant features by adding MMD as the domain alignment loss \cite{ding2018deep}. 
However, these methods only measuring the domain discrepancy neglect the difference among classes. 
Domain adaptation without class level domain discrepancy transfers source data to target domain leading to negative transfer.
Most recently, SimeNet \cite{pinheiro2018unsupervised} solves this problem by learning domain-invariant features and the categorical prototype representations. 
In the same way, CAN \cite{Kang_2019_CVPR} optimizes the network by considering the discrepancy of  intra-class domain and the inter-class domain.

\subsection{Generative Domain adaptation}

Another branch of unsupervised domain adaptation in DCNNs attempts to exploit a domain discriminator to address the domain confusion \cite{ganin2016domain,tzeng2017adversarial,zhang2018fully}.
Adversarial adaptation methods minimize the domain discrepancy with a domain discriminator. Due to the adversarial loss in the objective function, more transferable representations can be generated \cite{tzeng2017adversarial}. 
By adversarial adaptation methods, in theory, generator produces outputs identically distributed as source domain.
Meanwhile, with sufficient dataset, a GAN network with encode-decode construction can transfer the same set of samples in source domain to any random distribution of images in the target domain.
Moreover, adversarial losses with auxiliary loss can make sure that the learned function can transfer an individual source sample to the desired domain more effectively\cite{zhu2017unpaired}.
To further reduce domain discrepancy, \textit{Zhu et al.} introduces an identity loss to make sure the transfer can preserve semantic feature between the input and output.
The traditional methods like cycle loss and identity loss in cycleGAN are too restricted for domain adaption. Meanwhile, the previous process in domain adaption only focuses on the global transform.
Although it can reduce the distribution difference as cross domain, it destroys the class semantic feature in each sample.
It leads to that the classifier trained in this way cannot get the ideal result in the target domain.

Recently, GTA \cite{sankaranarayanan2018generate} solves this problem by modifying Auxiliary Classifier GAN(AC-GAN) and merging the class label and real/fake label. In addition to GTA, SymNets\cite{zhang2019domain} takes care of this problems by making the asymmetric design of source and target task classifiers sharing with them its layer neurons.
These methods apply the MMD in a single branch GAN structure and get an excellent result.
However, in their result, we find that when the number of samples in the source domain is insufficient compared with the target domain, the class accuracy can not get the best performance.
At the same time, another method to solve this problem employs the multi-branch general structure \cite{wang2019transferable}. 
And
CADA \cite{kurmi2019attending} proposes to find adaptable regions using some estimate of the discriminator.
It pays most attention to detect whether the part of the picture is transferable or not and does not take the class-level semantic feature into count.

Differently, we propose a dual generative cross-domain generation framework by interpolating two intermediate domains to bridge the domain gap.
Our proposed method leverages bi-directional cross-domain generators to make two intermediate domains and use additional target data with pseudo labels for learning two task-specific classifiers.
Our work is on the exploitation of building bi-directional generation network with two classifiers, which has not been fully explored in the literature.

\section{The Proposed Algorithm}

\subsection{Preliminary and Motivation}
Given the unlabeled target domain $X_{t} =  \{x_{t}\}^{M}_{j=1}$ and an auxiliary source domain $X_{s} = \{x_{s}\}^{N}_{i=1}$ with corresponding label $y_{s}$, the target of unsupervised domain adaption is to utilize the classifier trained in source domain to identify sample from target domain. 
Without loss of generality, the distribution of source data and target data are generally denoted as $x_{s}\sim p_{data}(x_{s})$ and $x_{t}\sim p_{data}(x_{t})$ respectively, and $ p_{data}(x_{t}) \neq p_{data}(x_{s})$. In order to solve the discrepancy of cross-domain, adversarial network based methods \cite{tzeng2017adversarial,bousmalis2017unsupervised,mancini2019adagraph} include two strategies: align two domains into a latent domain-invariant space \cite{lee2019sliced,chen2019progressive} and reconstruct one domain in another domain space \cite{zhang2019domain,roy2019unsupervised}.

The second strategy transferring source domain into target domain attempts to enable the distribution of source domain to be similar with that of target domain \cite{long2015learning,long2017deep}. 
Under this circumstance, the performance of the classifier trained in transferring source domain might be further promoted in target domain. 
However, due to the fact that samples from target domain have none label information, the boundary of class in target domain will be overlapping, which triggers overfitting and mismatching situation. 
Meanwhile unbalanced domain size may cause insufficient class-semantic feature when we make unsupervised domain adaption. Furthermore, the domain classifiers might find fewer counterparts to align due to cross-domain translations, rotations, or other transformations. Bi-Directional generative domain adaptation model with dual adversarial classifiers successfully remedy this weakness, enabling BDG to achieve excellent results.

Another strategy aligning the features of two domains in the same latent feature will suffer from the similar problems. 
They only employ a generator shared by the source domain and target domain \cite{long2015learning,tsai2018learning}. 
However, the classifier is just trained by the part of latent space belonging to the source domain. And the training process neglects the application of target domain. 
Thus, when the number of samples in the target domain is much more than that in the source domain, the well-trained classifier fails to have great performance in target domain. 
In order to take full advantage of the information of target domain, the bi-directional generation becomes a promising method.
In order to preserve semantic information in the picture, bi-directional generation based methods \cite{huang2018auggan} employ the identity loss with $L_{1}$ norm making a balance between generating samples and original samples. 
It can significantly reduce distribution discrepancy. However, these methods obviously lose the class-level semantic structure. 
Thus, we propose the Bi-Directional Generation for cross-domain learning with dual consistent classifiers to capture class-level semantic information and overcome the problems of overfitting and mismatching.

\subsection{Bi-Directional Cross-Domain Generation}

\begin{figure}
    \centering
    \includegraphics[width=0.48\textwidth]{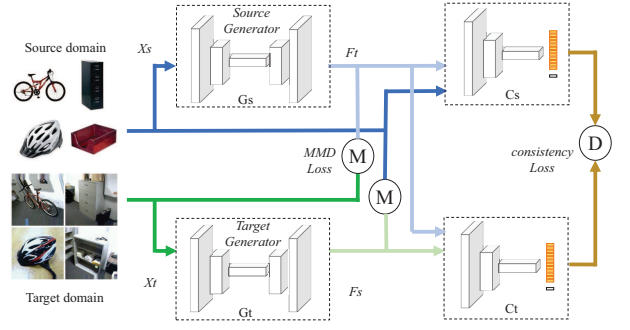}
    \caption{Overview of the proposed bi-directional generation network. It consists of two generators $G_s$ and $G_t$, two classifiers $C_s$ and $C_t$. $C_s$ is fed with $F_t$ and $X_t$; $C_t$ is fed with $F_s$ and $F_t$. $\mathcal{M}$ is used to calculate $\mathcal{L}_{MMD}$. $\mathcal{D}$ calculates discrepancy of the two prediction tensors to product $\mathcal{L}_{con}$.}
    \label{Overview grtaph}
\end{figure}

As illustrated in Figure \ref{Overview grtaph}, $X_{s}$, $X_{t}$ are source and target samples, respectively. 
We propose the cross-domain generators $G_s$, $G_t$ to transfer one domain input to the other domain distribution. 
Specifically, two generators are defined as $G_{s}:X_{s}\to X_{t}$ and $G_{t}:X_{t}\to X_{s}$, respectively. 
Given the source samples $X_s$, $G_{s}$ tries to generate $F_{t}$ that looks similar to target samples $X_{t}$. Similarly, With $X_t$, $G_{t}$ aims to generate $F_{s}$ which looks similar to $X_{s}$.

We define our bi-directional generation loss in source branch using the following formulation:
\renewcommand{\arraystretch}{1.2}
\begin{equation}
\begin{array}{l}
\label{eq: Source gan loss}
\mathcal{L}^s_{GAN}(X_s)=\mathcal{L}_{dis_s}+\mathcal{L}_{cls_s}\\
\mathcal{L}_{dis_s}=\mathbb{E}[\log C_s(X_s)] +\mathbb{E}[\log(1-C_s(G_s(X_s))) ]\\
\mathcal{L}_{cls_s}=\mathbb{E}[\log C_s(X_{s},Y_s)] +\mathbb{E}[\log C_s(G_s(X_s),Y_s)],
\end{array}
\end{equation}
where $\mathcal{L}_{dis_s}$ is the discrimination loss and the $\mathcal{L}_{cls_s}$ represents the classification loss. 
$C_{s}$ aims to classify the transferred samples $F_{t}$ and the original samples $X_{s}$, which means it can discriminate $X_s$ as real samples and $F_t$ as fake samples. 
Note that $F_t$ as a newly-augmented intermediate domain across source and target mitigates the domain gap. Meanwhile, $F_t$ generated from $G_s$ not only needs to be identified easily by $C_s$ but also makes $C_s$ have difficult in discriminating which domains it comes from.

Inspired by domain confusion \cite{tsai2018learning}, they forward two domain data to a fully-convolutional discriminator by using a cross-entropy loss for the two classes(\textit{i.e.} source and target). 
In the same way, $C_{t}$ aims to classify $F_{s}$ and $F_{t}$. 
We define our bi-directional generation loss in target branch using the following equation as:
\begin{equation}
\begin{array}{l}
\label{eq: target gan loss}
\mathcal{L}^t_{GAN}(X_s,X_t)=\mathcal{L}_{dis_t}+\mathcal{L}_{cls_t} \\
\mathcal{L}_{dis_t}=\mathbb{E}[\log C_t(G_s(X_s))] +\mathbb{E}[\log(1-C_t(G_t(X_t))) ] \\
\mathcal{L}_{cls_t}=\mathbb{E}[\log C_t(F_{t},Y_s)] +\mathbb{E}[\log C_t(F_s,\hat{Y}_t)],
\end{array}
\end{equation}

In order to align the two domains $X_t$ and $F_s$ on the target distributions, we employ a pre-trained classifier $C_{0}$ which is trained on the source domain without any auxiliary method. 
It produces the pseudo label of the target domain, $\hat{Y_{t}}=C_{0}(X_{t})$.
It is helpful to transfer data from target domain to source domain because $C_0$ preserves the class-level semantic information of source domain.
The discrimination part of $C_t$ is different from the general idea discriminating the real or fake sample. 
It distinguishes which domain the sample comes from. Specifically, it regards $F_t$ as samples from source domain and $F_s$ as samples belonging to target domain. Meanwhile, $F_t$ and $F_s$ generated from $G_s$, $G_t$ respectively enable $C_t$ to easily classify them and make $C_t$ have difficult in recognizing which domain they comes from. 
Thus, the proposed method can utilize more information from target domain.

\subsection{Class-wise Cross-Domain Alignment}

The Maximum Mean Discrepancy (MMD) \cite{gretton2012kernel} is a useful and popular non-parametric metric for measure discrepancy between cross domains.
The MMD loss is regarded as the discrepancy measure to compare distinct distributions, which computes the domain discrepancy between the sample means of the source and target data. 
Without this loss, we experimentally find that the performance of the proposed algorithm drops significantly.
Inspired by \cite{wang2019transferable,Kang_2019_CVPR}, we apply the class-level MMD loss as well. 
Our MMD loss includes two terms: \textit{global MMD} reducing the distance between two domains center point; and \textit{class MMD} reducing center point distance in each class between the source domain and target domain. 
We express the objective of MMD constraint in source and target domains as:
\begin{equation}
\label{eq:mmd 1}
\mathcal{L}^{s/t}_{MMD}=\mathcal{L}^{s/t}_{gMMD}+\frac{1}{C}\mathcal{L}^{s/t}_{cMMD},
\end{equation}
where $C$ is the class number. The global MMD in source branch is defined as:
\begin{small}
\begin{equation}
\label{eq:mmd 2}
\mathcal{L}^{s}_{gMMD}=\left\Vert {\frac{1}{n_{s}}\sum_{\mathclap{x_{s} \in X_{s}}}  G_{s}(x_{s})}   -  \frac{1}{n_{t}}\sum_{\mathclap{x_{t} \in X_{t}}} x_{t}  \right\Vert_{2},
\end{equation}
\end{small}
where $ n_{s}$, $n_{t}$ are the number of the source domain and target domain. 
It is easy to write similar equations for the target domain branch according to \eqref{eq:mmd 2}. 
And the class MMD in source branch is computed as:
\begin{small}
\begin{equation}
\label{eq:mmd 3}
\mathcal{L}^{s}_{cMMD}=\sum^{C}_{c}\left\Vert {\frac{1}{n_{s}^{c}}\sum_{\mathclap{x_{s} \in X^{c}_{s}}} G_{s}(x_{s})}  - \frac{1}{n_{t}^{c}}\sum_{\mathclap{x_{t} \in X^{c}_{t}}} x_{t}  \right\Vert_{2},
\end{equation}
\end{small}
where $n_{s}^{c}$, $n_{t}^{c}$ denote the number of its domain in class $c$. t is easy to write similar equations for the target domain branch according to \eqref{eq:mmd 3}. 
The difference between our work and others is that we calculate two MMD losses: the MMD loss between $F_t$ and $X_t$ as well as the MMD loss between $F_s$ and $X_s$. With pseudo labels of $X_t$ and $F_s$, we can easily calculate their $\mathcal{L}^t_{cMMD}$. This strategy aims to take full use of target domain and reduce the domain gap with two directions.

\subsection{Dual Consistent Classifiers}

Dual consistent classifiers refer to that no matter where the classifiers are trained, they should obtain very similar prediction outputs given the target data.
In our work, we employ two classifiers $C_t$ and $C_s$, which are trained by samples from different domains. However, in ideal situation, $C_t$ and $C_s$ should have similar prediction ability on target samples. 
So we introduce the $\mathcal{L}_{con}$ to regular the two target prediction outputs. 
The dual consistent classifiers loss is defined as:
\begin{equation}
\label{eq:dis loss}
\mathcal{L}_{con}=\left\Vert C_{t}(F_{s})-C_{s}(F_{s}) \right\Vert_{1},
\end{equation}
where $\mathcal{L}_{con}$ denotes the function measuring L1-norm between two probabilistic outputs computed from $C_t$ and $C_s$. 
This term indicates how the two classifiers agree on their predictions. 
Our goal is to make the two classifiers' prediction consistent so that it can reduce the domain gap further. Meanwhile, by utilizing this loss, the two classifiers can trigger each other. It is helpful to make the whole framework stable during whole training process.

\subsection{Overall Objective and Optimization}

To integrate cross-domain generation, discriminative class-wise alignment and dual consistent classifiers into one unified framework, we have our overall objective function as follows:
\begin{equation}
\label{eq:fullloss}
\mathcal{L}=\mathcal{L}^s_{GAN}+\mathcal{L}^t_{GAN}+\lambda(\mathcal{L}^s_{MMD}+\mathcal{L}^t_{MMD})+\gamma\mathcal{L}_{con},
\end{equation}
where $\lambda$, $\gamma$ control the relative importance of the two terms.
The $\mathcal{L}^s_{GAN}$ has been defined in equation \eqref{eq: Source gan loss} and the $\mathcal{L}^t_{GAN}$ can be calculated by equation \eqref{eq: target gan loss}. 
The $\mathcal{L}^{s/t}_{MMD}$ has been declared in equation \eqref{eq:mmd 1}.

To sum up the previous discussion, we need to train two classifiers which take inputs from the generator and minimize classifier loss, and the generators which try to mimic the the all objective loss. 
We optimize this objective in three steps.

\noindent\textbf{Step A} First, we train classifier $C_0$ to classify the source samples correctly and produce the pseudo label $\hat{Y_{t}}$ of the target domain $X_t$ and their alignment domain $F_s$. 
The pseudo label $\hat{Y_{t}}$ is utilized to calculate the $\mathcal{L}^{s/t}_{cMMD}$ and make a clear boundary of class in target domain, which is helpful to take full advantage of information from target domain. 
They are also beneficial to keep the balance of two branches' domain adaption, because the pseudo labels still maintain abundant domain-invariant representations to guide the $G_t$.
This step is crucial to implement next steps.
We train the networks $C_0$, whose structure is the same as the classifiers trained by next step, to minimize cross-entropy loss. 
The objective is as follows:
\begin{equation}
\label{eq:C0 loss}
\min_{C_0} \mathcal{L}(C_0,X_s)=\mathbb{E}[\log C_0(X_s)].
\end{equation}

\noindent\textbf{Step B} In this step, we train the classifiers ($C_s$, $C_t$) by fixing generators ($G_s$, $G_t$).
In order to take advantage of rich domain-invariant representations, we utilize data and their labels from both domains and train the final models for both branches.
The $C_s$ is fed by $X_s$ and $F_t$. Meanwhile $C_t$ is trained by $F_t$ with $y_s$, and $F_s$ with $\hat{y_t}$. Under this circumstance, the discrepancy between source domain and target domain will be reduced dramatically. 
In addition, the classifiers is not only to minimize the GAN losses but also they can also keep coincident in predicting target domain samples. 
To realize the this goal, we add a consistent loss on the loss of classifier.
The objective is as follows:
\begin{equation}
\label{eq:classifier loss}
\min_{C_s,C_t}\mathcal{L}^s_{GAN}+\mathcal{L}^t_{GAN} +\gamma\mathcal{L}_{con},
\end{equation}
 where these two classifiers can be optimized.

\noindent\textbf{Step C} We train the generators to minimize the full objective loss by using equation \eqref{eq:fullloss} for fixed classifiers. Unlike the general adversarial process in generator training, which minimizes the $\mathcal{L}_{dis_t}$ by regarding the fake samples as real samples, our method switch the both domain label at the same time. 
In detail, the $F_s$ drawn from target domain is regarded as source domain samples and the $F_t$ is considered as target domain samples.

\noindent{\textbf{Remarks}}: MCD \cite{saito2018maximum} employs two classifiers trained by source samples jointly with a feature generator.
However, through many experiments, we find that the total loss becomes diverge with epoch prolonging.
In our method, we adapt the discrepancy loss without classifier adversarial process.
We train the two classifiers $C_{s}$ processing $X_{s}$ as well as $F_{t}$ and $C_{t}$ with the input $F_{t}$ and $F_{s}$.

Obviously, the $C_{t}$, $C_{s}$ are different classifiers. Compared with the \cite{saito2018maximum} which shares generator between both domain, we employ two independent generators $G_{t}$, $G_{s}$ to implement bi-direction domain adaption. 
We add the MMD loss taking the class-level structure information into count, into the total loss to to replace the discrepancy loss in \cite{saito2018maximum}. 
Meanwhile we adapt discrepancy loss as consistent loss, which is optimized by classifiers and generators. 
Thus, we do not maximize the discrepancy loss when we train the classifiers. 
In other words, reducing MMD loss aims to guide the generators on how to map the source domain to the target domain.

\section{Experiments and Results}

We evaluate BDG method by the standard benchmarks including Office-31 and Office-Home, compared with state of the art domain adaption methods.

\subsection{Datasets and Experimental Setup}

\textbf{Office-31} \quad \cite{saenko2010adapting}, a standard benchmark for visual domain adaptation, contains 4,652 images and 31 categories from three distinct domains, \textit{i.e.}, images collected from the 1) Amazon website (Amazon domain), 2) web camera (Webcam domain), and 3) digital SLR camera (DSLR domain) under different settings, respectively. 
The dataset is imbalanced across domains, with 2,817 images in Amazon domain, 795 images in Webcam domain, and 498 images in DSLR domain. 
We follow the standard evaluation protocols for unsupervised domain adaptation \cite{ganin2016domain,long2015learning}.

\noindent\textbf{Office-Home} \quad \cite{venkateswara2017deep} is a more challenging dataset for domain adaptation evaluation. 
It consists of around 15,500 images in total from 65 categories of everyday objects in office and home scenes. There are four significantly different domains: Artistic images (Ar),  Clip Art (Cl), Product images (Pr), and Real-World images (Rw). The images of these domains have substantially different appearances and backgrounds. 
The the number of categories is much larger than that of Office-31, making it more difficult to transfer across domains. 
We evaluate all methods on all 12 adaptation tasks.

\subsection{Implementation Details}

The hyper-parameters $\lambda$,$\gamma$ in the equation\eqref{eq:fullloss} are selected as 1 throughout all experiments. 
We use ResNet-50\cite{he2016deep} models pre-trained on the ImageNet dataset \cite{russakovsky2015imagenet} as the backbone and we remove its last FC layer. 
And We fine-tune all convolutional and pooling layers and apply back-propagation to train the classifiers and generators. 
There optimizer of the classifiers $C_s$, $C_t$ is mini-batch stochastic gradient descent (SGD) with the momentum of 0.9. 
At the same time, we use adaptive moment estimation (Adam) to train the generators $G_s$, $G_t$, as in \cite{salimans2016improved}. 
And the learning rate is set to $5.0\times10^{-4}$ in all experiments. 
We report the accuracy result after 20,000 iterations.

To empirically verify the merit of our proposed model, We compare with both conventional and the state of the art transfer learning methods including:
\textbf{RTN} \cite{long2016unsupervised},
\textbf{ADDA} \cite{tzeng2017adversarial},
\textbf{JAN} \cite{long2017deep},
\textbf{SimeNet} \cite{pinheiro2018unsupervised},
\textbf{GTA} \cite{sankaranarayanan2018generate},
\textbf{TADA} \cite{wang2019transferable},
\textbf{STA} \cite{liu2019separate},
\textbf{SymNets} \cite{zhang2019domain},
\textbf{SAFN} \cite{xu2019larger}. 
All of them have been introduced in related works section.
we also compare with other domain adaption methods:
\textbf{ResNet-50} \cite{he2016deep} directly exploits the classification model trained on the source domain to classify target samples; \textbf{DANN} \cite{ganin2016domain} designs a domain regularizer to calculated the $\mathcal{H}$-divergence; \textbf{MADA} \cite{pei2018multi} takes multi-mode structures to get the suitable alignment of different data distributions based on multiple domain discriminators; 
\textbf{DSR} \cite{ijcai2019-285} reconstructs the semantic latent variables and domain latent variables by employing a variational auto-encoder.

\renewcommand{\arraystretch}{1.5}
\begin{table*}[ht]
\caption{Accuracy (\%) on Office-31 for unsupervised domain adaption (ResNet)}
\label{tab:table1}
\centering
\resizebox{\textwidth}{!}{
\begin{tabular}{lllllllllllll}
\toprule
Method & ResNet-50 & RTN & DANN & ADDA & JAN & MADA & SAFN & SimNet & GTA & SymNets & TADA & BDG \\
\midrule
A$\to$W & 68.4$\pm$0.2 & 84.5$\pm$0.2 & 82.0$\pm$0.4 & 86.2$\pm$0.5 & 85.4$\pm$0.3 & 90.0$\pm$0.1 & 88.8$\pm$0.4 & 88.6$\pm$0.5 & 89.5$\pm$0.5 & 90.8$\pm$0.1 & \textbf{94.3$\pm$0.3} & 93.6$\pm$0.4 \\
D$\to$W & 96.7$\pm$0.1 & 96.8$\pm$0.1 & 96.9$\pm$0.2 & 96.2$\pm$0.3 & 97.4$\pm$0.2 & 97.4$\pm$0.1 & 98.4$\pm$0.0 & 98.2$\pm$0.2 & 97.9$\pm$0.3 & 98.8$\pm$0.3 & 98.7$\pm$0.1 & \textbf{99.0$\pm$0.1} \\
W$\to$D & 99.3$\pm$0.1 & 99.4$\pm$0.1 & 99.1$\pm$0.1 & 98.4$\pm$0.3 & 99.8$\pm$0.2 & 99.6$\pm$0.1 & 99.8$\pm$0.0 & 99.7$\pm$0.2 & 99.8$\pm$0.4 & \textbf{100.0$\pm$0.0} & 99.8$\pm$0.2 & \textbf{100$\pm$0.0} \\
A$\to$D & 68.9$\pm$0.2 & 77.5$\pm$0.3 & 79.7$\pm$0.4 & 77.8$\pm$0.3 & 84.7$\pm$0.3 & 87.8$\pm$0.2 & 87.7$\pm$1.3 & 85.3$\pm$0.3 & 87.7$\pm$0.5 & \textbf{93.9$\pm$0.5} & 91.6$\pm$0.3 & 93.6$\pm$0.3 \\
D$\to$A & 62.5$\pm$0.3 & 66.2$\pm$0.2 & 68.2$\pm$0.4 & 69.5$\pm$0.4 & 68.6$\pm$0.3 & 70.3$\pm$0.3 & 69.8$\pm$0.4 & 73.4$\pm$0.8 & 72.8$\pm$0.3 & \textbf{74.6$\pm$0.6} & 72.9$\pm$0.2 & 73.2$\pm$0.2 \\
W$\to$A & 60.7$\pm$0.3 & 64.8$\pm$0.3 & 67.4$\pm$0.5 & 68.9$\pm$0.5 & 70.0$\pm$0.4 & 66.4$\pm$0.3 & 69.7$\pm$0.2 & 71.6$\pm$0.6 & 71.4$\pm$0.4 & 72.5$\pm$0.5 & \textbf{73.0$\pm$0.3} & 72.0$\pm$0.1 \\
Avg & 76.1 & 81.6 & 82.2 & 82.9 & 84.3 & 85.2 & 85.7 & 86.2 & 86.5 & 88.4 & 88.4 & \textbf{88.5}\\
\bottomrule
\end{tabular}}
\end{table*}

\renewcommand{\arraystretch}{1.3}
\begin{table*}[ht]
\caption{Accuracy (\%) on Office-Home for unsupervised domain adaption (ResNet)}
\label{tab:table2}
\resizebox{\textwidth}{!}{
\begin{tabular}{llllllllllllll}
\toprule
Method & Ar$\to$Cl & Ar$\to$Pr & Ar$\to$Rw & Cl$\to$Ar & Cl$\to$Pr & Cl$\to$Rw & Pr$\to$Ar & Pr$\to$Cl & Pr$\to$Rw & Rw$\to$Ar & Rw$\to$Cl & Rw$\to$Pr & Avg \\
\midrule
ResNet-50 & 34.9 & 50.0 & 58.0 & 37.4 & 41.9 & 46.2 & 38.5 & 31.2 & 60.4 & 53.9 & 41.2 & 59.9 & 46.1 \\
DANN  & 45.6 & 59.3 & 70.1 & 47.0 & 58.5 & 60.9 & 46.1 & 43.7 & 68.5 & 63.2 & 51.8 & 76.8 & 57.6 \\
JAN & 45.9 & 61.2 & 68.9 & 50.4 & 59.7 & 61.0 & 45.8 & 43.4 & 70.3 & 63.9 & 52.4 & 76.8 & 58.3 \\
DSR & \textbf{53.4} & 71.6 & 77.4 & 57.1 & 66.8 & 69.3 & 56.7 & 49.2 & 75.7 & 68.0 & 54.0 & 79.5 & 64.9 \\
SymNets & 47.7 & 72.9 & 78.5 & 64.2 & 71.3 & \textbf{74.2} & 64.2 & 48.8 & 79.5 & 74.5 & 52.6 & 82.7 & 67.6 \\
TADA & 53.1 & 72.3 & 77.2 & 59.1 & 71.2 & 72.1 & 59.7 & \textbf{53.1} & 78.4 & 72.4 & \textbf{60.0} & 82.9 & 67.6 \\
BDG & 51.5 & \textbf{73.4} & \textbf{78.7} & \textbf{65.3} & \textbf{71.5} & 73.7 & \textbf{65.1} & 49.7 & \textbf{81.1} & \textbf{74.6} & 55.1 & \textbf{84.8} & \textbf{68.7}\\
\bottomrule
\end{tabular}}
\end{table*}

\subsection{Comparison Results}

The classification accuracy on the Office-31 dataset for unsupervised domain adaptation based on ResNet-50 are shown in Table \ref{tab:table1}. 
For a fair comparison, the results of all baselines are directly reported from their original papers wherever available. 
To be honest, the results on D$\to$A is little lower than existing methods. 
However, we get reasonable results in almost task, \textit{i.e.} A$\to$W and A$\to$D. Moreover, the performance in task W$\to$D and D$\to$W is over all methods.
Taking all into count, the BDG model outperforms all compared methods on mean accuracy.

According to Table \ref{tab:table2}, the BDG approach overpasses the compared methods on all transfer tasks on Office-Home. 
Moreover, it improves their accuracy significantly in many tasks, even though this dataset has abundant categories.
Compared with TADA\cite{wang2019transferable}, the BDG improves more than 6 percents in some transfer learning tasks, such as Cl$\to$Ar, Pr$\to$Ar.
It illustrates that BDG yields larger improvements on such difficult transfer learning tasks. 
And it also suggests that BDG can maintain more class-semantic feature during effective domain adaptation. Moreover, in the office-home dataset existing unbalanced samples number problem, the classifier can implement sufficient adversarial process, which makes the two classifiers be more consistent. 
Meanwhile, the large dataset is helpful to calculate $\mathcal{L}^{s/t}_{MMD}$, especially $\mathcal{L}^{s/t}_{cMMD}$, because mismatching problem can be avoided,\textit{i.e.}, some class in $F_t$ or $F_s$ is blank.

\subsection{Empirical Analysis}

\subsubsection{t-SNE visualization}

\begin{figure*}
\centering  
\subfigure[ResNet]{
\label{a}
\hspace{-4mm}\includegraphics[width=0.23\textwidth]{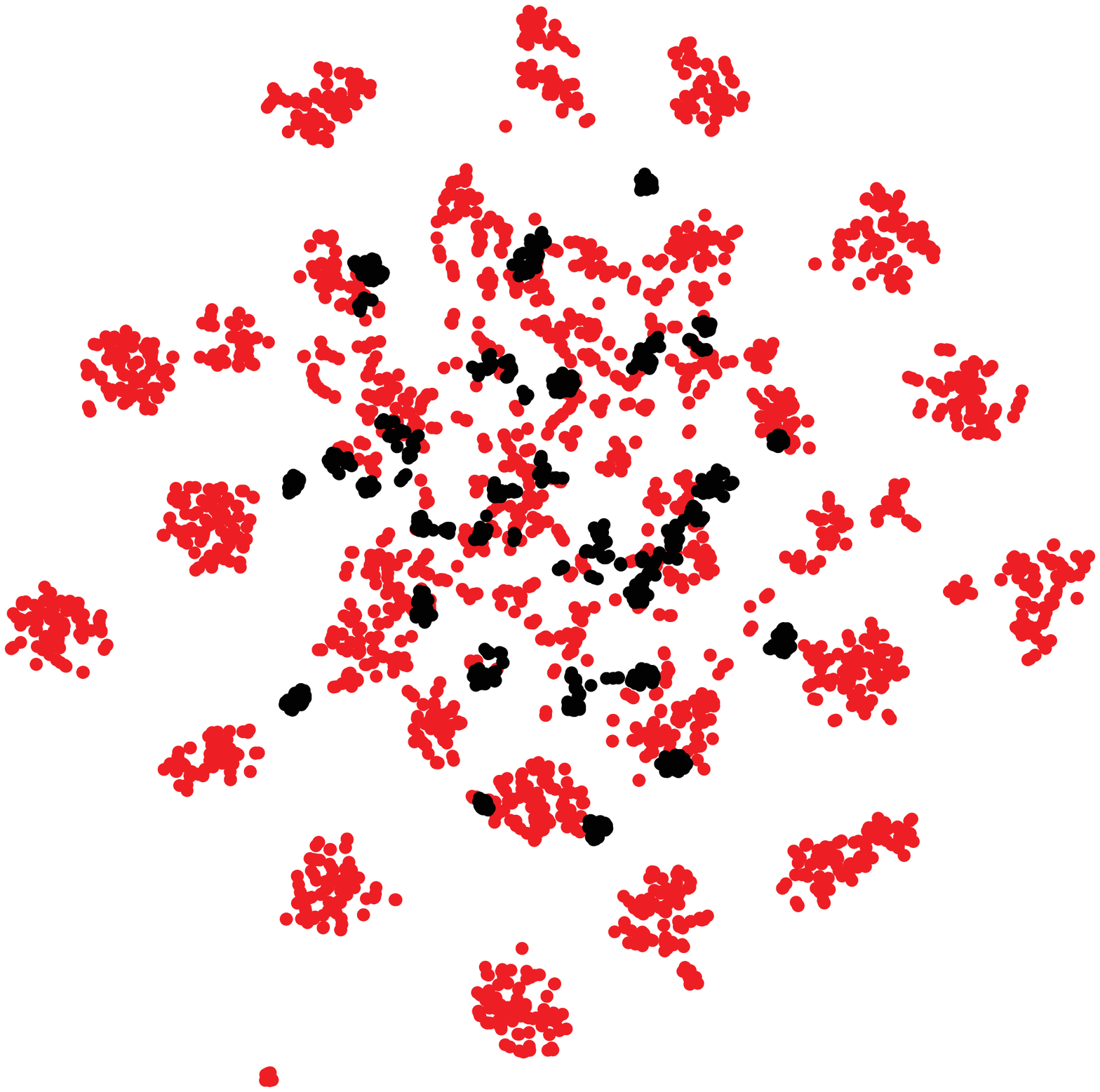}}
\subfigure[DAN]{
\label{b}
\includegraphics[width=0.235\textwidth]{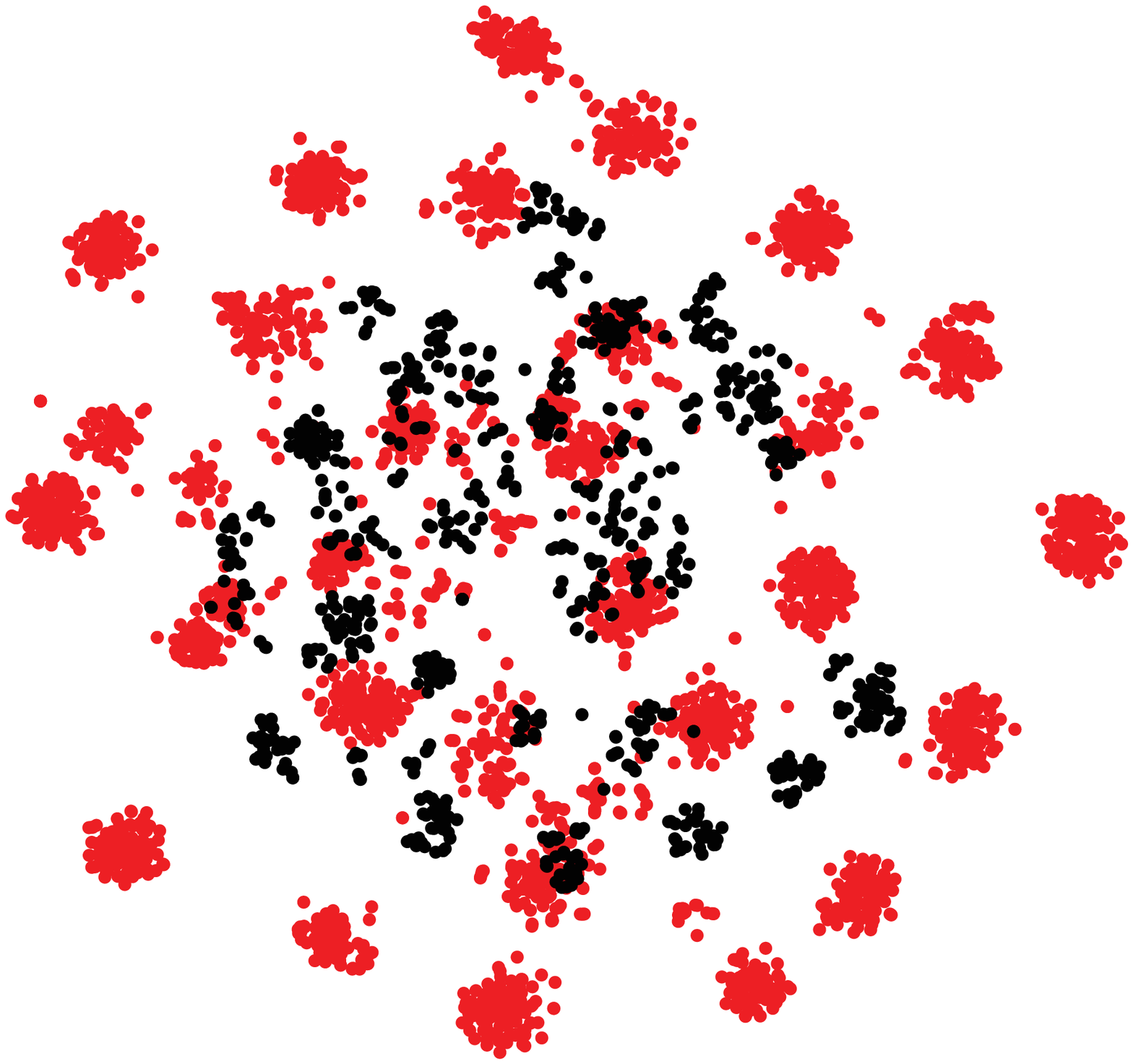}}
\subfigure[RevGrad]{
\label{c}
\includegraphics[width=0.235\textwidth]{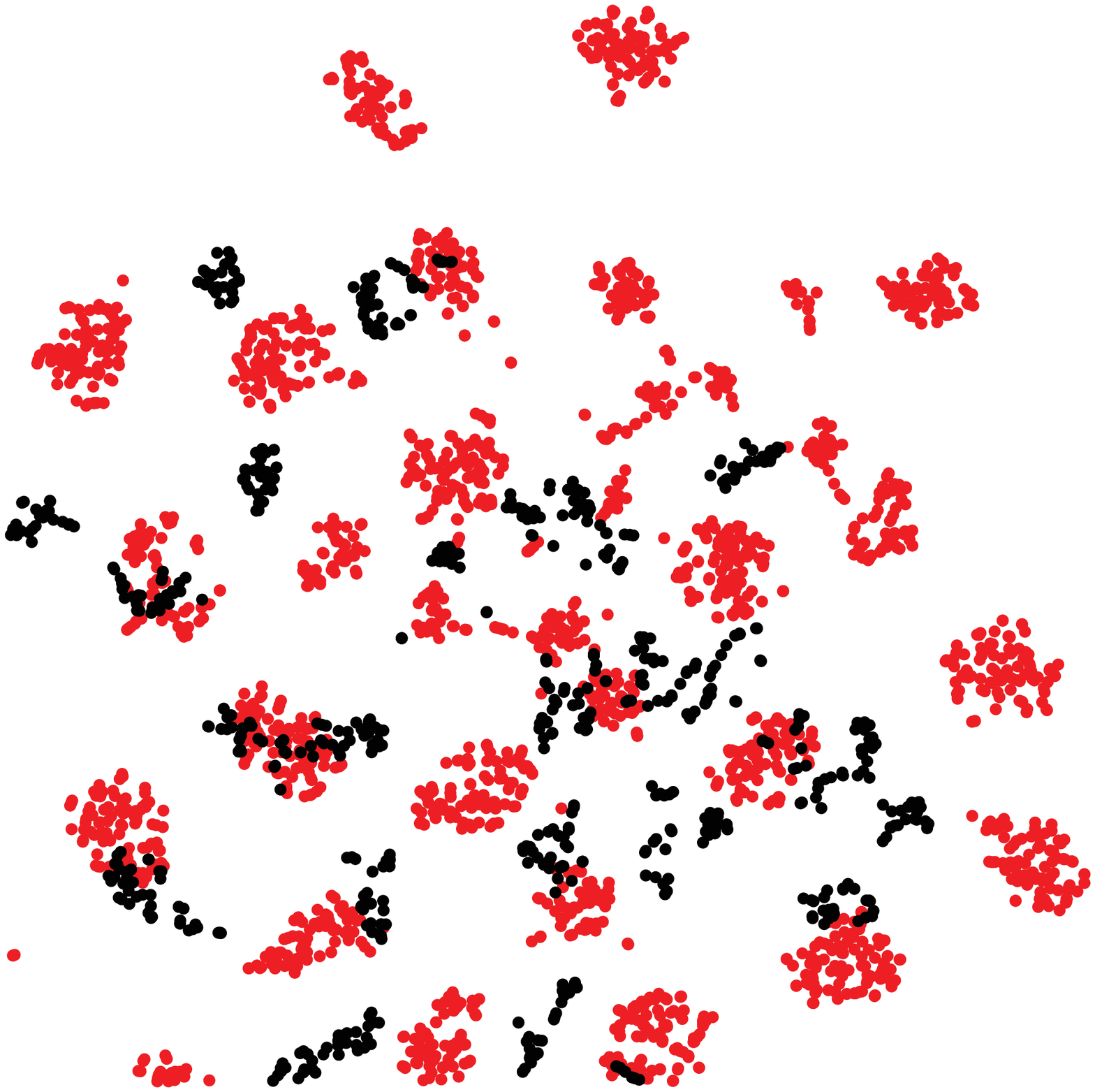}}
\subfigure[BDG]{
\label{d}
\includegraphics[width=0.235\textwidth]{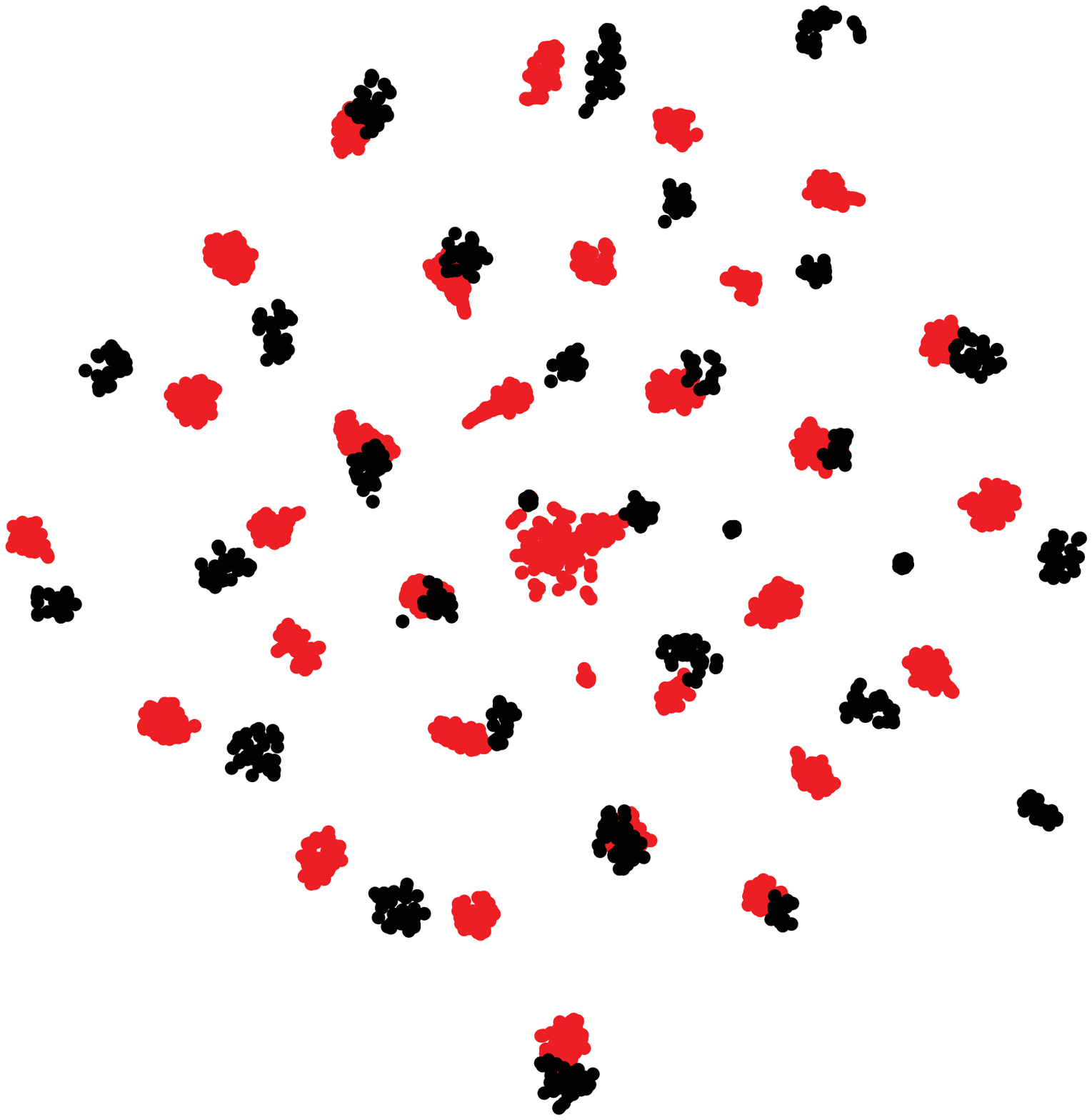}}
\caption{Visualization with t-SNE for different adaptation methods (bested viewed in color).\textbf{a}: t-SNE of ResNet. \textbf{b}: DAN.\textbf{c}: RevGrad.\textbf{d}: BDG.
The input activations of the last FC layer are used for the computation of t-SNE. The results are on Office-31 task $Amazon \to Webcam$.}
\label{t-SNE}
\end{figure*}

To further understand the alignment of distribution, we visualize the features in 2D-space using the network activations of the FC layer from task $Amazon \to webcam$ (31 classes) learned by ResNet, DAN, RevGrad and BDG, respectively using t-SNE embeddings \cite{donahue2014decaf}. 
The representations generated by BDG (Figure \ref{t-SNE}) form exactly 31 clusters with clear boundaries. 
Compared to ResNet, DAN and RevGrad, as expected, our t-SNE figure demonstrates the closer distance between the same classes in different domain. 
This shows that our BDG generate more discriminative features for both domain and confirms our improvement in Tables \ref{tab:table1} \& \ref{tab:table2}.

\subsubsection{Parameter analysis} 

\begin{figure*}[ht]
\centering 
\subfigure[Webcam$\to$Amazon (Office-31)]{
\label{a}
\includegraphics[width=0.32\textwidth]{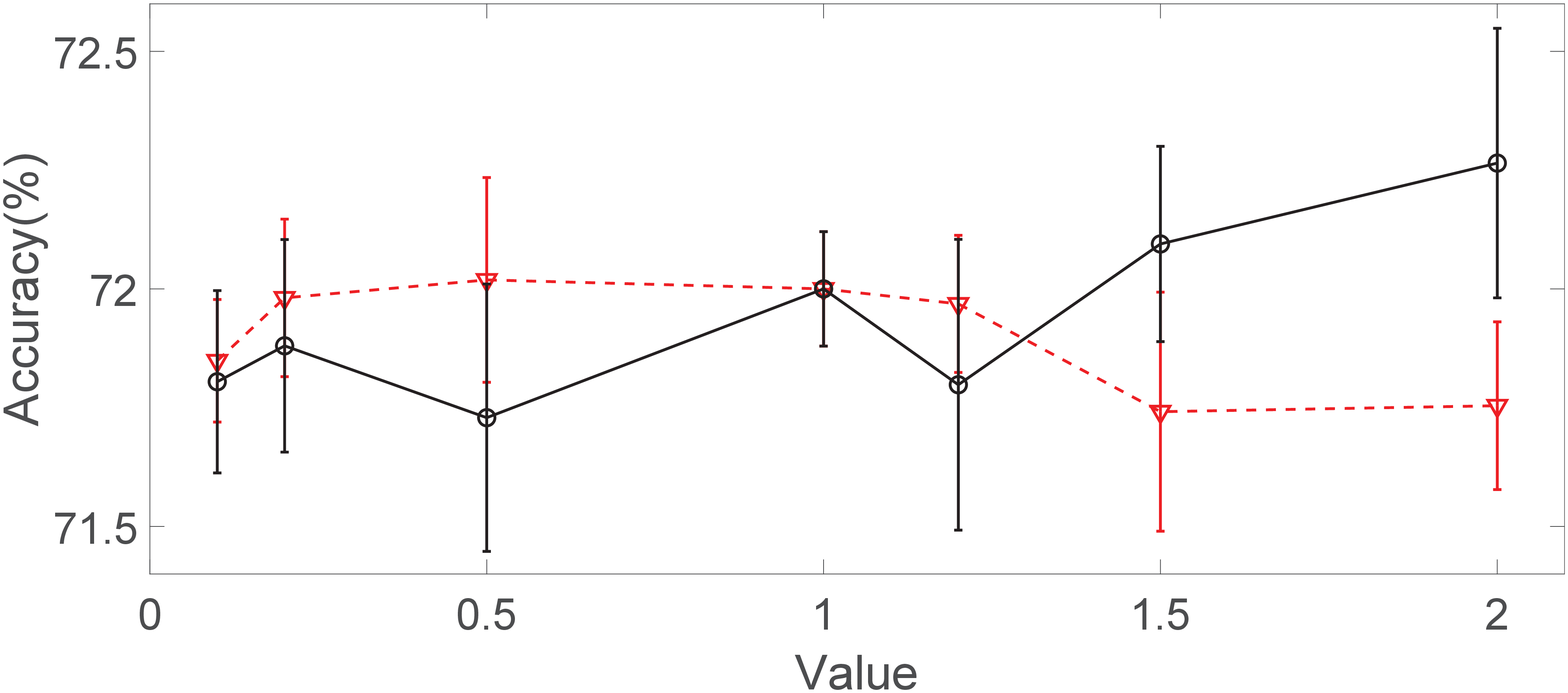}}
\subfigure[Realworld$\to$Product (Office-home)]{
\label{b}
\includegraphics[width=0.32\textwidth]{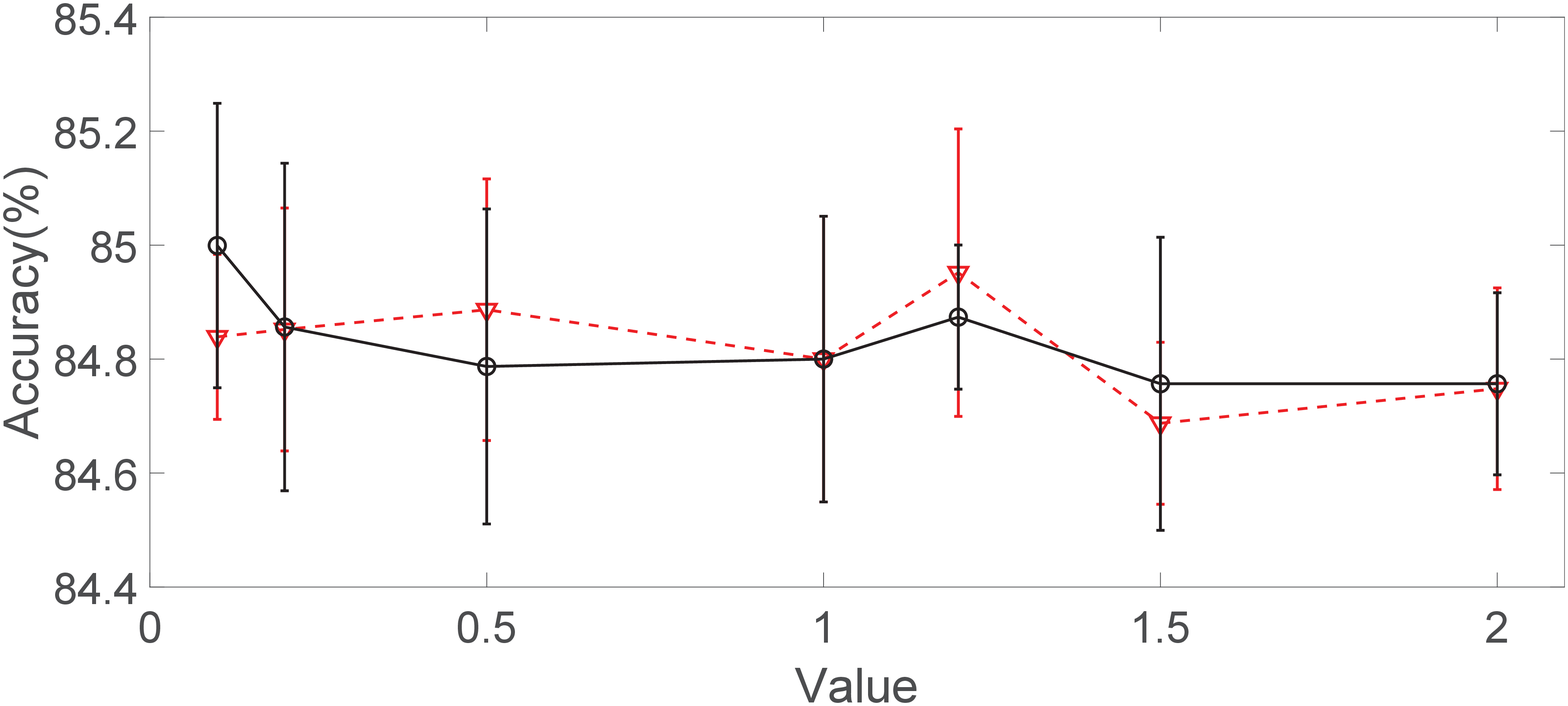}}
\subfigure[Art$\to$Realworld (Office-home)]{
\label{c}
\includegraphics[width=0.32\textwidth]{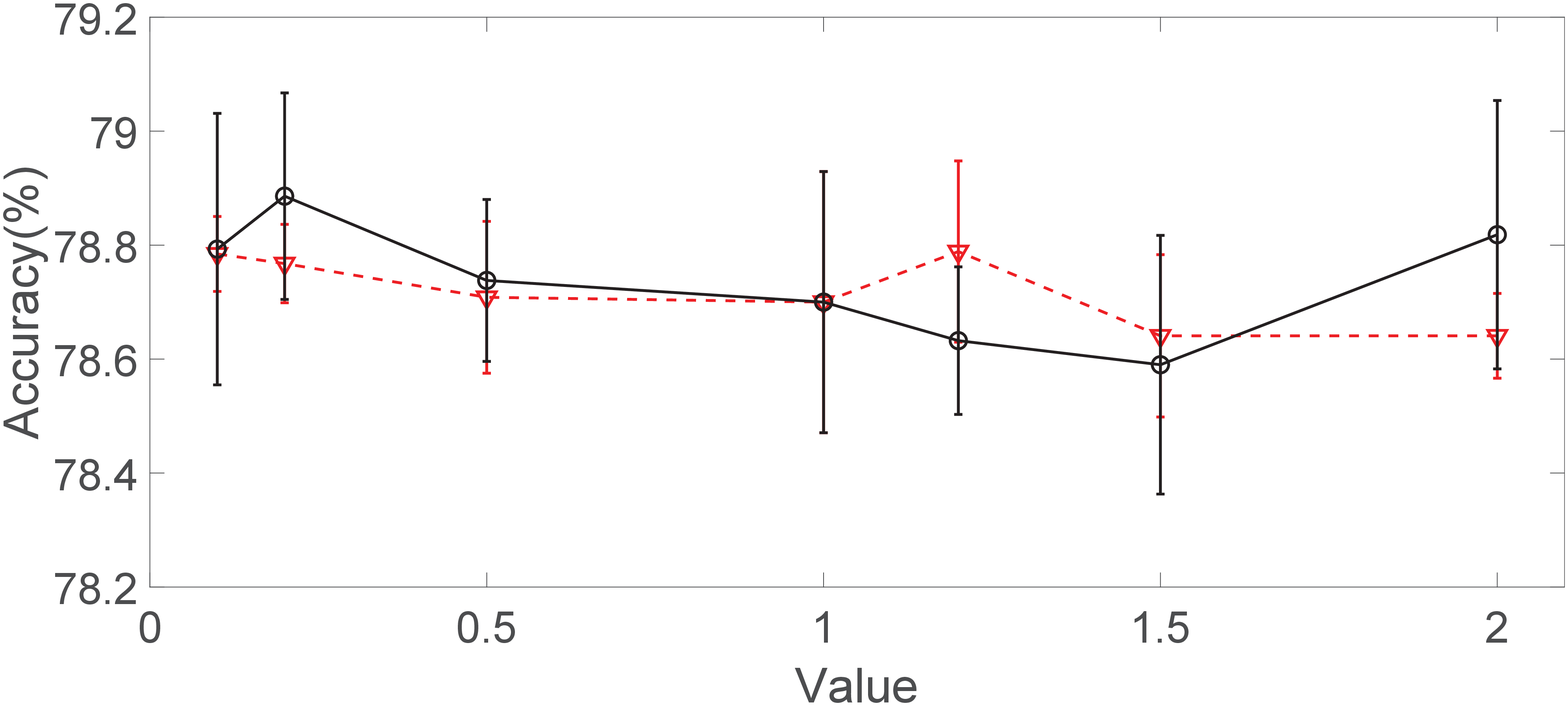}}
\caption{Parameter study of our algorithm: averaged accuracy with parameter $\lambda$,$\gamma$ learned by (a) webcam$\to$amazon (Office-31), (b) Art$\to$Realworld (Office-Home), (c) Art$\to$Realworld  (Office-Home)(black:$\lambda$, red:$\gamma$)}
\label{Parameter study}
\end{figure*}

We conduct experiments to investigate the sensitivity of our method to the balance parameters $\lambda$, $\gamma$. We use control variations method to test two super-parameters, whose value are selected at the range of $[0.1, 0.2, 0.5, 1 ,1.2, 1.5 ,2]$. For example, when we change the value of $\lambda$, $\gamma$ is fixed to 1. From Figure \ref{Parameter study}, as $\lambda$ go up, the average accuracy of office-31 dataset increases while accuracy of office-home decreases gradually. 
At the same time, as $\gamma$ increase, the average accuracy of every dataset raise and then go down gradually. 
In conclusion, hyper-parameters is quite effective, but the optimal value in each case is different. 
We fix $\lambda=1$ and $\gamma=1$ in the other experiments because its result is more stable than others.

\renewcommand{\arraystretch}{1.5}
\begin{table}[!ht]
\caption{The effect of MMD loss and discrepancy loss. The mean accuracy of Office-home and office-31 are reported.}
\label{tab:table3}
\resizebox{0.48\textwidth}{!}{
\begin{tabular}{lllllll}
\toprule
dataset & Variant1 & Variant2 & Variant3 & Variant4 & Variant5 & BDG \\
\midrule
Office-31 & 81.1 & 87.4 & 82.9 & 82.5 & 88.3 & 88.5 \\
Office-Home & 61.0 & 67.8 & 63.2 & 63.5 & 68.2 & 68.7 \\
\bottomrule
\end{tabular}}
\end{table}

\subsubsection{Ablation studies} We compare our method(``BDG'') with other variants, to verify the effect of $\mathcal{L}^{s/t}_{MMD}$ and $\mathcal{L}^{s/t}_{con}$. 
We adopt the single directional cross-domain generation method translating target domain to the source domain without $\mathcal{L}^{s/t}_{MMD}$ as \textbf{variants 1}. Compared to \textbf{variant 1}, we design another method named \textbf{variant 2} which includes the loss function of MMD. 
\textbf{variant 3} is the bi-directional Cross-Domain Generation method with dual classifiers. 
In addition, we also design the \textbf{variant 4} consisting of $\mathcal{L}^{s/t}_{GAN}$ and $\mathcal{L}_{con}$. 
Moreover, we also propose the \textbf{variant 5} including $\mathcal{L}^{s/t}_{GAN}$ and $\mathcal{L}^{s/t}_{MMD}$ to do comparison with other variants. 
It can be seen that combining $\mathcal{L}^{s/t}_{MMD}$ and $\mathcal{L}_{con}$ improves the adaption performance.
In detail, the mean accuracy of Variant 2(87.4\% in the office-31 dataset) increase significantly, compared with the result(81.1\%) of variant 1. 
The main reason is that the $\mathcal{L}^{s/t}_{MMD}$ reduces the domain discrepancy and maintains the class-level information. 
The similar situation also happens in the office-home dataset. 
When we compare the variant 3,4,5 and BDG in Table \ref{tab:table3}, it can be seen that the result of variant 4 in two datasets is close to the result of variant 3. 
In detail, the result of variant 4 in the office-home dataset is over only 0.3\% than variant 3.
However, when the loss combines the $\mathcal{L}^{s/t}_{MMD}$ and $\mathcal{L}_{con}$, the performance improves. 
So we can get some conclusions according to Table \ref{tab:table3}. 
First of all, it can be proved that $\mathcal{L}^{s/t}_{MMD}$ takes an essential role in our method.
Second, it also can demonstrate $\mathcal{L}_{con}$ is an applicable term in our method. 
The last but not the least, bi-directional cross-domain generation method does work better than single branch structure method.

\begin{figure}[!ht]
    \centering
\includegraphics[width=0.49\textwidth]{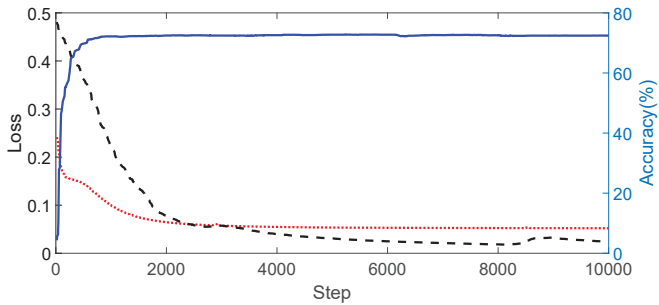}
    \caption{The curve of $\mathcal{L}_{MMD}$ (red), $\mathcal{L}_{discrepancy}$ (black) and accuracy(blue) during training on task Webcam $\to$ Amazon of the office-home dataset.}
    \label{MMD+acc+dis curve}
\end{figure}

\subsubsection{Convergence Analysis} 

The Figure \ref{MMD+acc+dis curve} shows the whole training process of our BDG method on task Rw$\to$Pr from office-home dataset. 
We notice that the total loss becomes diverge with epoch prolonging if the full loss include adversarial loss. Therefore, they can get an excellent result in the middle process but their final result may be much weaker than the best. We solve these problems by adding the $\mathcal{L}^{s/t}_{MMD}$ and $\mathcal{L}_{con}$. 
The Figure\ref{MMD+acc+dis curve} also reports that accuracy gradually increases when the $\mathcal{L}^{s/t}_{MMD}$ and $\mathcal{L}_{con}$ continuously decrease. 
It approves that the training process avoids becoming diverge with epoch prolonging due to the effect of two terms.

\section{Conclusion}

In this paper, we propose a bi-directional generative cross-domain generation framework to explore domain adaptation in an unsupervised manner.
Previous adversarial domain adaption approaches with shared generator neglect class features by interpolating two intermediate domains to bridge the domain gap.
We add the MMD loss and consistent loss into loss function to implement bi-directional Cross-Domain Generation method. 
The proposed method includes MMD loss which takes full use of target domain and reduces the domain gap with two directions, and the consistent loss which utilizes the task-specific classifiers to align the source and target features. 
Experiments on two standard domain adaption benchmarks verifies the advanced effectiveness of our algorithm which has better performance than other state-of-the-art domain adaptation models.

\newpage
{
\small
\bibliographystyle{aaai}
\bibliography{ref}
}

\end{document}